 \documentclass[pmlr,twocolumn]{jmlr} 

\usepackage{booktabs}
\usepackage[load-configurations=version-1]{siunitx} 

 \usepackage{float}
 
\usepackage{bbold}
\newcommand\numberthis{\addtocounter{equation}{1}\tag{\theequation}}

\theorembodyfont{\upshape}
\theoremheaderfont{\scshape}
\theorempostheader{:}
\theoremsep{\newline}

\usepackage[normalem]{ulem}

\jmlrvolume{ML4H Extended Abstract Arxiv Index}
\jmlryear{2020}
\jmlrsubmitted{2020}
\jmlrpublished{}
\jmlrworkshop{Machine Learning for Health (ML4H) 2020}

\title[Estimating County-Level COVID-19 Exponential Growth Rates]{Estimating County-Level COVID-19 Exponential Growth Rates Using Generalized Random Forests}

\makeatletter
\renewcommand*{\thanks}[1]{%
  \footnotemark
  \protected@xdef\@thanks{\@thanks
    \protect\footnotetext[\arabic{footnote}]{#1}}%
}
\makeatother

\makeatletter
\newcommand{\printfnsymbol}[1]{%
  \textsuperscript{\@fnsymbol{#1}}%
}
\makeatother

\newcommand\extrafootertext[1]{%
    \bgroup
    \renewcommand\thefootnote{\fnsymbol{footnote}}%
    \renewcommand\thempfootnote{\fnsymbol{mpfootnote}}%
    \footnotetext[0]{#1}%
    \egroup
}

\author{%
\Name{Zhaowei She}\printfnsymbol{1} \Email{zshe3@gatech.edu}\\
\addr Georgia Institute of Technology
\AND
\Name{Zilong Wang}\printfnsymbol{1} \Email{ zwang937@gatech.edu}\\
\addr Georgia Institute of Technology
\AND
\Name{Turgay Ayer} \Email{ayer@isye.gatech.edu}\\
\addr Georgia Institute of Technology
\AND
\Name{Asmae Toumi} \Email{atoumi@mgh.harvard.edu}\\
\addr  Massachusetts General Hospital
\AND
\Name{Jagpreet Chhatwal} \Email{jagchhatwal@mgh.harvard.edu}\\
\addr  Massachusetts General Hospital, Harvard Medical School
}

\begin{document}

\maketitle

\begin{abstract}
Rapid and accurate detection of community outbreaks is critical to address the threat of resurgent waves of COVID-19. A practical challenge in outbreak detection is balancing  accuracy vs. speed. In particular, while estimation accuracy improves with longer
fitting windows, speed degrades. This paper presents a machine learning framework to balance this tradeoff using generalized random forests (GRF), and applies it to detect county level COVID-19 outbreaks. This algorithm chooses an adaptive fitting window size for each county based on relevant features affecting the disease spread, such as changes in social distancing policies. Experiment results show that our method outperforms any non-adaptive window size choices in 7-day ahead COVID-19 outbreak case number predictions.

\end{abstract}

\section{Introduction}


Early and accurate detection of community outbreaks is critical to address the threat of resurgent waves of COVID-19. Specifically, an epidemic outbreak is confirmed when its incident cases are estimated to grow exponentially. Furthermore, the potential impact of an outbreak is also measured by its exponential growth rate, as higher rates indicate more rapid disease spread. Last but not least, for a fixed epidemiology model (e.g. SIR, SEIR), there is an one-to-one correspondence between the exponential growth rate of an epidemic outbreak and its basic reproduction number $R_0$, a common measure of intensity of epidemic outbreaks \citep{lipsitch2003transmission}. Therefore, the exponential growth rate of an epidemic outbreak's incident cases is the most important "model-free" parameter to estimate for detecting the outbreak \citep{chowell2003sars}.\par
It remains an epidemiological challenge to obtain accurate exponential growth rate estimates for disease outbreaks \citep{ma2014estimating}. Specifically, it is difficult to choose the fitting window size for the exponential growth rate estimation to balance the speed and accuracy of outbreak detection. On one hand, a longer fitting window is preferable as larger sample size would reduce variance of the exponential growth rate estimates of outbreaks. On the other hand, shorter fitting windows are better at detecting early-stage outbreaks, especially if these outbreaks were driven by recent policy changes such as school reopening. In the current practice, this fitting window size is treated as a hyperparameter that is either directly specified by the user (c.f. \cite{Mel}) or determined by some cross validation methods (c.f. \cite{chowell2007comparative}).\par

This paper develops a machine learning framework that balances the speed-accuracy tradeoff of outbreak detection via dedicated feature engineering and GRF (c.f. \cite{athey2019generalized}), and apply it to detect county-level COVID-19 outbreaks. Specifically, the algorithm chooses an adaptive fitting window size for each county based on a rich set of features that affect the disease spread, such as face mask mandates, social distancing policies, the CDC’s Social Vulnerability Index, changes in tests performed and rate of positive tests. Furthermore, for counties with insufficient data to capture the recent policy changes, the algorithm pools together all relevant incident case growth trends across U.S. counties and throughout the COVID-19 pandemic history to adjust for these policy changes.\par



\section{Background}

\subsection{Exponential Growth Model and Exponential Growth Rate}

During an epidemic outbreak, incident case number at county $c\in C\subset \mathbb{N}$ on day $t\in T \subset \mathbb{N}$, i,e. $I_{t,c}$, is governed by an exponential growth model, 
\begin{align*}
   &\ \ \ \ \ \ \ \ \  I_{t,c} = I_{0,c}\exp{\{rt\}}\\
& \iff  ln(I_{t,c}) = ln(I_{0,c}) + r t. \numberthis \label{EG}
\end{align*}
\citep{ma2014estimating}.\footnote{See Appendix \ref{apdA} for the epidemiological definition of incident case number and how we compute $I_t$.} Here $r$ is the incident cases exponential growth rate of outbreaks for all counties $c\in C$, while $I_{0,c}$ captures the initial incident case numbers at the beginning of the exponential case growth at county $c$. To obtain the most recent exponential growth rate of county-level COVID-19 incident case number, we estimate the instantaneous counterpart of (\ref{EG}),
\begin{equation}\label{EEG}
    ln(I_{t,c}) = \alpha_{t,c} + r_{t,c} t + \varepsilon_{t,c},
\end{equation}
where the dependent variable is the log-linearized incident case number $ln(I_{t,c})$; independent variable is day $t$. The parameter of interest is the COVID-19 incident case exponential growth rate of county $c$ at day $t$, $r_{t,c}$. The intercept term, $\alpha_{t,c}$, is a parameter capturing the log-linearized initial incident case number of county $c$ at the beginning of the outbreak. $\varepsilon_{t,c}$ is the error term.\par

Notably, the exponential growth rate $r_{t,c}$ in (\ref{EEG}) varies in both day $t$ and county $c$. In other words, we are interested in estimating the instantaneous county-level exponential growth rate of COVID-19 incident cases. Specifically, since COVID-19 related regulations are changing every day in the U.S. (c.f. \cite{CUSP}), the county-level exponential growth rates, affected by these policies, also change from day to day. Therefore, in order to detect recent outbreaks, we need to estimate the most recent incident cases exponential growth rate. This instantaneous exponential growth rate $r_{t,c}$, similar to the instantaneous reproduction number commonly used in the epidemiology literature (c.f. \cite{fraser2007estimating,cori2013new}), captures the projected exponential growth rate of incident cases in county $c$ should the future COVID-19 regulations remain the same as those in day $t$.\par

\subsection{Relevant Features Affecting COVID-19 Disease Spread}

The exponential growth rate of COVID-19 incident cases can be affected by many factors ranging from day-to-day changes in COVID-19 regulations to difference in population density and healthcare resources between counties. Hence, in order to estimate the instantaneous county-level exponential growth rate $r_{t,c}$ defined in (\ref{EEG}), we need to control for these day-level and county-level heterogeneity. Provided that we have relevant features $X_{t,c}\in \mathbb{R}^m$ that captures these aforementioned factors affecting the instantaneous county-level exponential growth rate $r_{t,c}$,\footnote{Refer to Appendix \ref{apdC} for the list of relevant features used in this study.} we can identify the conditional average partial treatment effect $r_{t,c}(X_{t,c}):=\mathbb{E}[r_{t,c}|X_{t,c}]$ as defined by \cite{wooldridge2010econometric}. When data of these relevant features affecting COVID-19 disease spread is available, we can rewrite (\ref{EEG}) as
\begin{equation}\label{FEEG}
    ln(I_{t,c}) = \alpha_{t,c}(X_{t,c}) + r_{t,c}(X_{t,c}) t + \varepsilon_{t,c}
\end{equation}
following the common ``redundant" assumption (c.f. \cite{wooldridge2010econometric}), i.e.

\begin{assumption}\label{A1}$\forall t\in T, \forall c\in C, \forall X_{t,c} \in \mathbb{R}^m$
\begin{align*}
  \mathbb{E}[ln(I_{t,c})|t,\alpha_{t,c},r_{t,c},X_{t,c}]
  = \mathbb{E}[ln(I_{t,c})|t,\alpha_{t,c},r_{t,c}]
\end{align*}
\end{assumption}

\section{Model Estimation}

This section discusses how we estimate the exponential growth rate of COVID-19 incident cases $r_{t,c}(X_{t,c})$ defined in (\ref{FEEG}). Specifically, we first formulate the estimation problem in \S\ref{PF} and then present our estimation algorithm in \S\ref{EA}. 

\subsection{Problem Formulation}\label{PF}

First, we need the following ``unconfoudness" assumption (c.f. \cite{rosenbaum1983central}):
\begin{assumption}\label{A2} $\forall t\in T, \forall X_{t,c}\in \mathbb{R}^m$
\begin{equation*}
   \mathbb{E}[ln(I_{t,c})-\alpha_{t,c}(X_{t,c}) - r_{t,c}(X_{t,c}) t|t,X_{t,c}]=0.
\end{equation*}
\end{assumption}
That is, day $t$ is independent of all unobservable heterogeneity conditional on the feature vector $X_{t,c}$. Under Assumption \ref{A2}, we can derive the following moment equations for (\ref{FEEG}): $\forall X_{t,c}\in \mathbb{R}^m$,
\begin{align*}
    &\mathbb{E}[\Gamma(t)(ln(I_{t,c})-\alpha_{t,c}(X_{t,c}) - r_{t,c}(X_{t,c}) t)|X_{t,c}]\\
    &=0, \numberthis \label{me}
\end{align*}
through the instrumental variable $\Gamma(t):=[\mathbb{1}_{0}(t);\mathbb{1}_{1}(t);\mathbb{1}_{2}(t);\dots]$, where $\mathbb{1}_{s}(t)$ is a Boolean function equal 1 if $t=s$.\par 

However, we note that the moment equations (\ref{me}) cannot identify the exponential growth rate $r_{t,c}(X_{t,c})$. Specifically, the moment equation system (\ref{me}) is underidentified as the number of unknown parameters $\{\alpha_{t,c}(X_{t,c}),r_{t,c}(X_{t,c})\}_{t\in T, c\in C}$ is exactly two times the number of moment equations. To address this problem, we assume that the exponential growth rate of day $t^*$ equal the local average causal response (c.f. \citet{abadie2003semiparametric,angrist1995two}) from day $t^*-1$ to day $t^*$, i.e.
\begin{assumption}\label{A3}\\ $\forall t^*\in T \backslash \{0\}, \forall c\in C, \forall X_{t^*,c}\in \mathbb{R}^m$
\begin{align*}
  & r_{t^*,c}(X_{t^*,c})\\
  &=\frac{\mathbb{E}[ln(I_{t,c})|\mathbb{1}_{t^*}(t)=1,X_{t^*,c}]}{\mathbb{E}[t|\mathbb{1}_{t^*}(t)=1,X_{t^*,c}]-\mathbb{E}[t|\mathbb{1}_{t^*-1}(t)=1,X_{t^*,c}]}\\
  &-\frac{\mathbb{E}[ln(I_{t,c})|\mathbb{1}_{t^*-1}(t)=1,X_{t^*,c}]}{\mathbb{E}[t|\mathbb{1}_{t^*}(t)=1,X_{t^*,c}]-\mathbb{E}[t|\mathbb{1}_{t^*-1}(t)=1,X_{t^*,c}]}
\end{align*}
\end{assumption}
Parameters $\{\alpha_{t,c}(X_{t,c}),r_{t,c}(X_{t,c})\}_{t\in T, c\in C}$ are exactly identified by the moment equations (\ref{me}) under Assumption \ref{A3}.




\subsection{Estimation Algorithms}\label{EA}
Given the above assumptions, the exponential growth function of COVID-19 incident cases (\ref{FEEG}) can be estimated using the GRF algorithm by \cite{athey2019generalized}. Specifically, Assumption \ref{A3} implies that the exponential growth rate of a county $c$ on day $t^*$ is identified by a targeted data ``block" $\{(I_{t^*,c},X_{t^*,c}), (I_{t^*-1,c},X_{t^*-1,c})\}$. Hence, when estimating $r_{t^*,c}(X_{t^*,c})$, we can partition the panel dataset $\{(I_{t,c},X_{t,c})\}_{t\in T, c\in C}$ into $|C|\times \left \lfloor{\frac{t^*}{2}}\right \rfloor $ data blocks, and pool those data blocks ``similar" to the targeted block to construct an adaptive window size for this estimation. Particularly, the similarity measure is provided by the GRF algorithm. In Appendix \ref{apdB}, Algorithm \ref{alg:block} provides the pseudocode to construct these data blocks. Algorithm \ref{alg:grf} explains how we feed these data blocks into the GRF algorithm to obtain the estimates of interest.



\section{Performance Evaluation}

To benchmark our method against non-adaptive window size choices, we compare the Mean Absolute Percentage Error (MAPE) of these methods' 7-day ahead predictions. Specifically, we use the NYTimes COVID-19 Dataset (c.f. \cite{NYTimes2020}) as the source of daily reported cases per county. Additional results for performance evaluation are available at Appendix \ref{apdD}. 


As shown in Figure \ref{fig:MAPE4} and Table \ref{tab:4}, our method outperforms methods with fixed 2-, 4-, 8- or 16-day fitting window sizes. Specifically, Figure \ref{fig:MAPE4} shows that our method provides a uniformly better performance roughly 100 days after the first recorded COVID-19 case in the NYTimes COVID-19 Dataset, when there were enough historical data for the GRF algorithm to conduct meaningful partition. Furthermore, Table \ref{tab:4} demonstrates that even if the early-day MAPEs are included in the comparison, our method still has the best median MAPE among the 4 methods. Last but not least, we note that when only comparing the performance of non-adaptive window size choices, there are no obvious best choice. While choosing shorter fitting window sizes could in general lead to lower median MAPEs (c.f. Table \ref{tab:4}), it was still frequently outperformed by longer fitting window size choices (c.f. Figure \ref{fig:MAPE4}). 






\begin{figure}[ht]
\floatconts
  {fig:MAPE4}
  {\caption{MAPE Plot (4-day Moving Average)}}
  {\includegraphics[width=\linewidth]{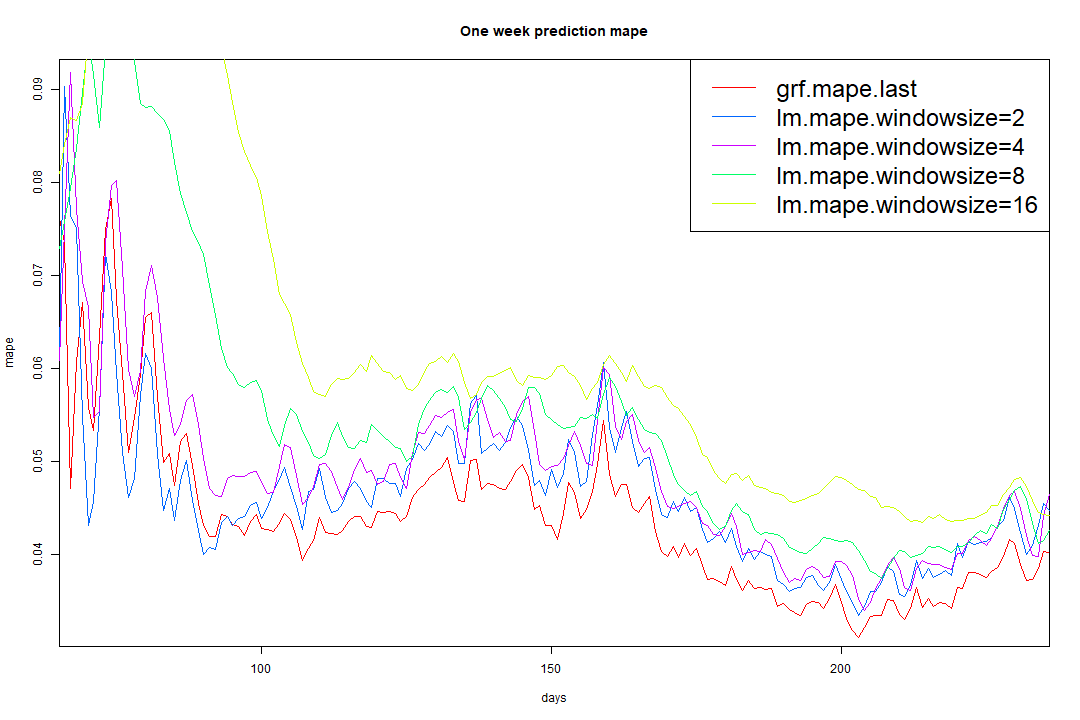}}
\end{figure}
\begin{table}[ht]
\floatconts
  {tab:4}%
  {\caption{Median RMSE and MAPE (4-day Moving Average)}}%
  {\begin{tabular}{ll}
  \toprule
  \bfseries Method &  \bfseries MAPE\\
  \midrule
  OLS.wsize=16 & 0.0585 \\
  OLS.wsize=8  & 0.0529 \\
  OLS.wsize=4  & 0.0483\\
  OLS.wsize=2  & 0.0455\\
  \textbf{GRF}  & \textbf{0.0423}\\
  \bottomrule
  \end{tabular}}
\end{table}

\section{Conclusion and Future Work}
In this work, we developed a novel framework that allows GRF to adequately balance the the speed-accuracy tradeoff in COVID-19 outbreak detection. This estimation framework can be readily extended to other epidemiology problems where fitting window size impacts model performance.



  


\bibliography{jmlr-sample}

\newpage

\appendix

\section{Pseudocode for Estimation Algorithms}\label{apdB}

\begin{algorithm}[H]\small
\floatconts
  {alg:block}%
  {\caption{Block Transformation}}%
{%
\KwIn{$\{(I_{t,c},X_{t,c})\}_{t\in T, c\in C}$}
\KwOut{$\{Feature[t,c]\}_{t\in T, c\in C}$, $\{Y[t,c,1], Y[t,c,0]\}_{t\in T, c\in C}$}
\For{$c \leftarrow 0$ \KwTo $|C|$}{
\For{$t \leftarrow 1$ \KwTo $|T|$}{
  \begin{enumerate*}
    \item Normalize the incident case numbers for each block 
    \begin{align*}
        Y[t,c,1] &\leftarrow ln(I_{t,c})-ln(I_{t-1,c})\\
        Y[t,c,0] &\leftarrow 0
    \end{align*}
    \item Generate the initial rough OLS estimates for each block
    \begin{align*}
        Dep &\leftarrow \{Y[t,c,0],Y[t,c,1]\}\\
        Ind &\leftarrow  \{0,1\}\\
      r_{ols}, \alpha_{ols} &\leftarrow OLS(Dep \sim Ind )
    \end{align*}
    \item Append the feature data for each block
    \begin{align*}
      Feature[t,c] &\leftarrow \{ X_{t,c}, r_{ols}, \alpha_{ols}, I_{t-1,c} \}
    \end{align*}
  \end{enumerate*}
}}
}%
\end{algorithm}

\begin{algorithm}
\floatconts
  {alg:grf}%
  {\caption{GRF Training for Day $t^*$}}%
{%
\KwIn{$\{Feature[t,c]\}_{t\in [0,1,\dots,t^*], c\in C}$, $\{Y[t,c,1], Y[t,c,0]\}_{t\in [0,1,\dots,t^*], c\in C}$}
\KwOut{$\{r_{t^*,c}(X_{t^*,c})\}_{c\in C}$}
  \begin{enumerate*}
    \item Compute the congruence classes\\ for day $t^*$ 
    \begin{align*}
        [t^*]&\leftarrow \{z\leq t^*| t^* \equiv z\ (\textrm{mod}\ 2)\}\\
        [t^*-1]&\leftarrow \{z\leq t^*| t^*-1 \equiv z\ (\textrm{mod}\ 2)\}
    \end{align*}
    \item Assign the feature variable
    \\ for day $t^*$ 
    $$X \leftarrow \{x_t\}_{t\in [0,1,\dots,t^*]}$$
    where
    \begin{equation}
    x_t = \begin{cases}
                Feature[t,c] &\text{if } t \in [t^*]\\
                Feature[t+1,c] &\text{if }  t \in [t^*-1]
             \end{cases} 
\end{equation} 
    \item Assign the outcome variable
    \\ for day $t^*$ 
    $$Y \leftarrow \{y_t\}_{t\in [0,1,\dots,t^*]}$$
    where
    \begin{equation}
    y_t = \begin{cases}
                Y[t,c,1] &\text{if } t \in [t^*]\\
                Y[t,c,0] &\text{if }  t \in [t^*-1]
             \end{cases} 
\end{equation} 
    \item Assign the treatment variable
    \\ for day $t^*$ 
    $$W \leftarrow \{w_t\}_{t\in [0,1,\dots,t^*]}$$
    where
    \begin{equation}
    w_t = \begin{cases}
                1 &\text{if } t \in [t^*]\\
                0 &\text{if }  t \in [t^*-1]
             \end{cases} 
\end{equation} 
    \item Feed (X,Y,W) into the GRF algorithm
    $$\{r_{t^*,c}(X_{t^*,c})\}_{c\in C} \leftarrow GRF(X,Y,W)$$
  \end{enumerate*}
}%
\end{algorithm}

\newpage 

\section{Epidemiological Definitions}\label{apdA}

For this work, where we wish to provide useful estimates of case trajectories for epidemiologists and decision makers in the government, 
we ideally wish to measure the growth rate of \textbf{active} case numbers 
\begin{equation}
    A_{t} := Y_{t} - D_{t} - R_{t}, \ t\geq 0 \text{ days },
\end{equation} 
where $Y$ is the cumulative number of cases so far, $D$ is the cumulative number of deaths, 
and $R$ is the cumulative number of recovered cases. 
This intuitively captures the number of still "infectious" cases, as we can assume for COVID-19 that those who have recovered or died are no longer capable of infecting others. 
Unfortunately, as the number of recovered cases $R_{t}$ are no longer reported, we are unable to directly calculate the number of active cases. While other works (c.f. \cite{Mel}) approximate the number of recovered cases with 
\begin{equation}
    R_{t} \approx \begin{cases}
                Y_{t-22} - D_{t-22} &\text{, if } t \geq 22\\
                Y_{t} - D_{t} &\text{, if }  0 \leq t < 22.
             \end{cases} 
\end{equation}
Here the underlying assumption is that those who were infected but not dead after 22 days have recovered. 
However, upon testing this assumption at the county-level, we find that this assumption does not hold 
as there will be some days where this newly approximated active case number becomes negative.

We hence rely upon a commonly used proxy: the \textbf{incident} case count, $I_{t}$, defined as 
\begin{equation}
    I_{t} := \begin{cases}
                Y_{t} - Y_{t-22} &\text{, if } t \geq 22\\
                Y_{t} &\text{, if }  0 \leq t < 22,
             \end{cases} \label{eq:3}
\end{equation} 
which is the number of new cases within a 22 day time period and is a useful proxy for infectious cases.

\section{Feature Data}\label{apdC}

All feature data we used, i.e. $\{X_{t,c}\}_{t\in T, c\in C}$, are publicly available online. In this section, we briefly describe what these datasets are and how we incorporated them in our work.


\subsection{2019 US Census Gazetteer Files}
The 2019 United States Census Gazetteer Files (c.f. \cite{USCensus2019}) were used to obtain the geographic locations (latitude-longitude centroids) of each county officially registered by the United States Census Bureau. This provides a spatial feature space for the GRF to further split upon.   

\subsection{Centers for Disease Control and Prevention Social Vulnerability Index 2018 Database}
The Social Vulnerability Index (SVI) database (c.f. \cite{CDCSVI2018}) is a compilation of socio-economic factors such as unemployment rate, poverty rate, education attainment level etc. at the county level. These features are also included in our methodology.

\subsection{COVID-19 US state policy database (CUSP)}
The CUSP database (c.f. \cite{CUSP}) tracks when each state implemented and ended policies such as mask mandates, lockdowns, economic policies in response to the COVID-19 pandemic. As such, it is a vector of features capturing daily policy changes in each state. As these policies is state-wide, we naturally extend them to the respective county level.     

\subsection{The COVID Tracking Project }
From the COVID Tracking Project (c.f. \cite{COVIDTracking}), we obtained the he daily numbers of PCR, Antibody and Antigen tests performed and their positivity rates at each state. These are used as features in our framework as well.

\section{Additional Results for Performance Evaluation}\label{apdD}

\subsection{MAPE Plots}

\begin{figure}[ht]
\floatconts
  {fig:MAPE3}
  {\caption{MAPE Plot (3-day Moving Average)}}
  {\includegraphics[width=\linewidth]{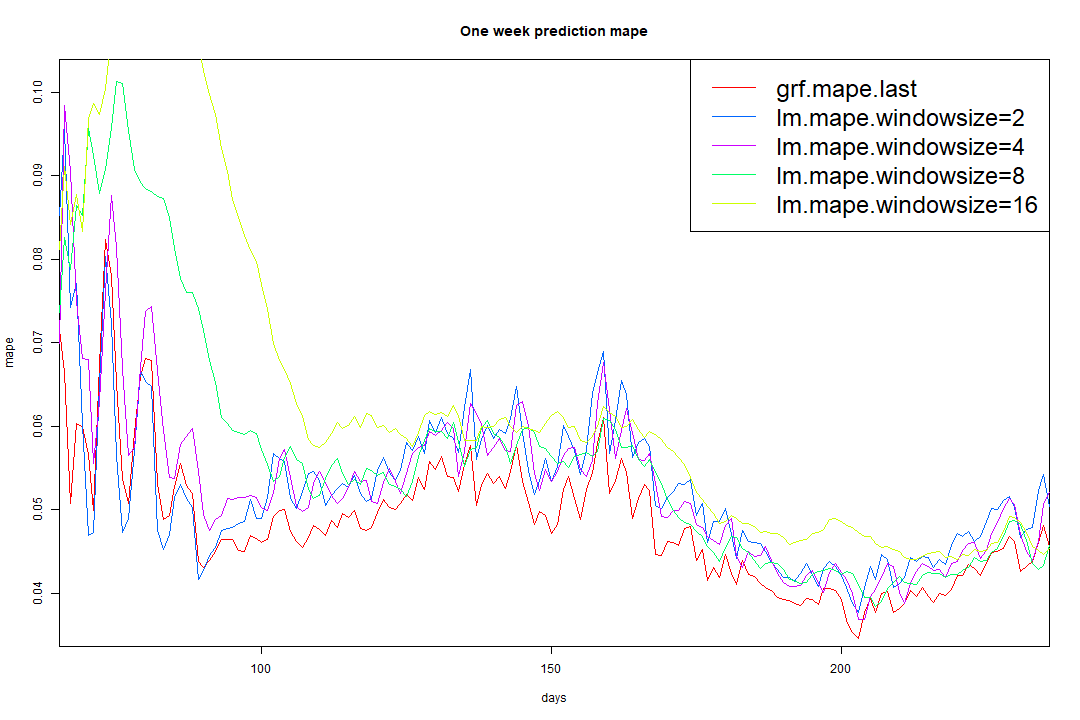}}
\end{figure}

\begin{figure}[ht]
\floatconts
  {fig:MAPE4L}
  {\caption{MAPE Plot (4-day Moving Average)}}
  {\includegraphics[width=\linewidth]{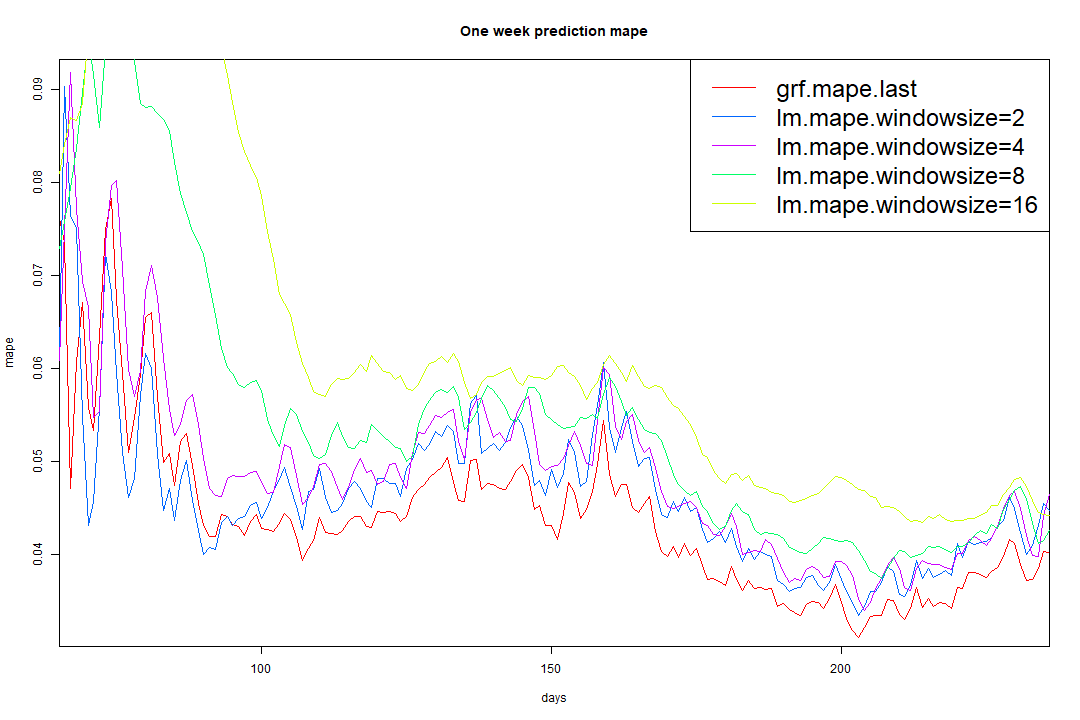}}
\end{figure}

\begin{figure}[ht]
\floatconts
  {fig:MAPE5}
  {\caption{MAPE Plot (5-day Moving Average)}}
  {\includegraphics[width=\linewidth]{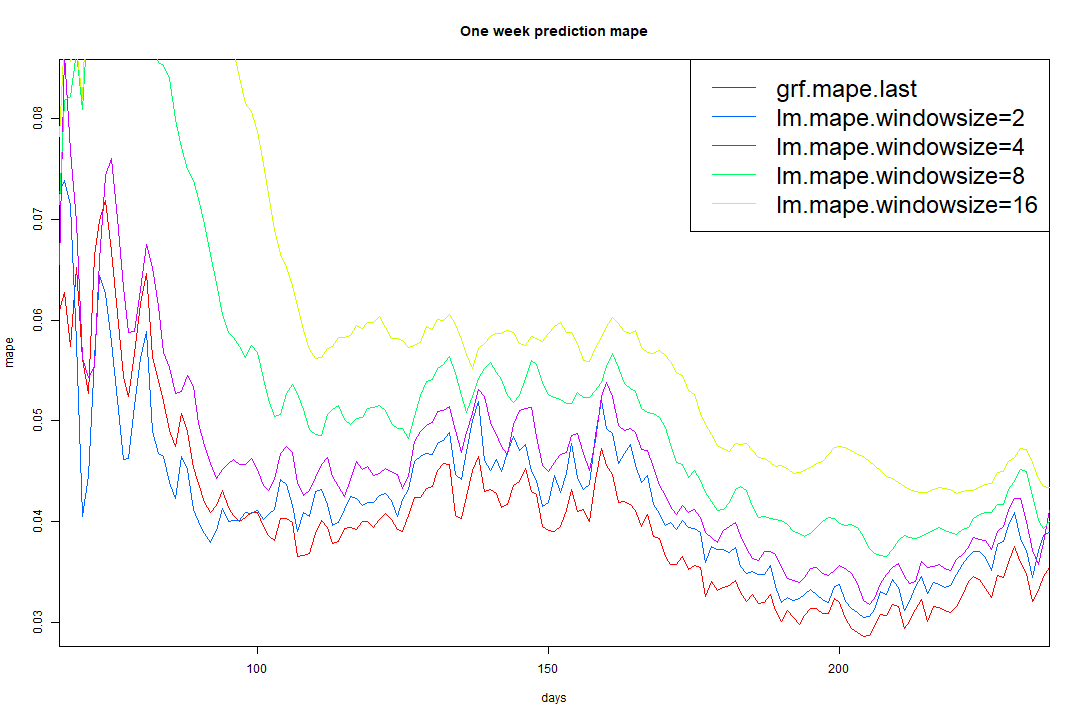}}
\end{figure}

\begin{figure}[ht]
\floatconts
  {fig:MAPE6}
  {\caption{MAPE Plot (6-day Moving Average)}}
  {\includegraphics[width=\linewidth]{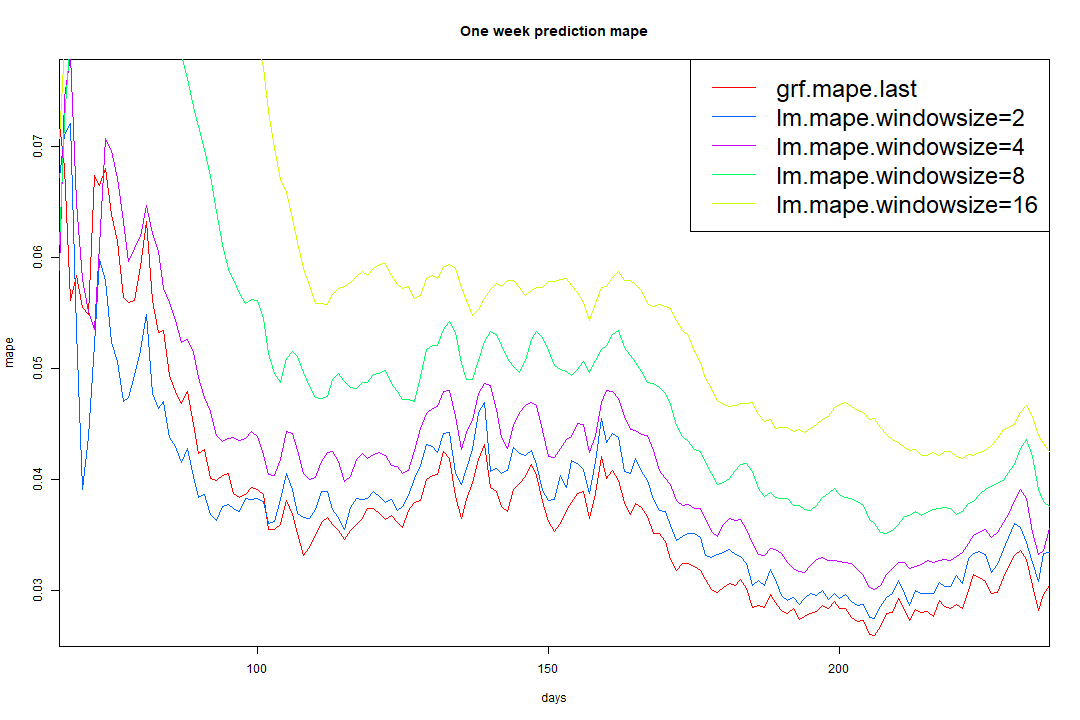}}
\end{figure}

\begin{figure}[ht]
\floatconts
  {fig:MAPE7}
  {\caption{MAPE Plot (7-day Moving Average)}}
  {\includegraphics[width=\linewidth]{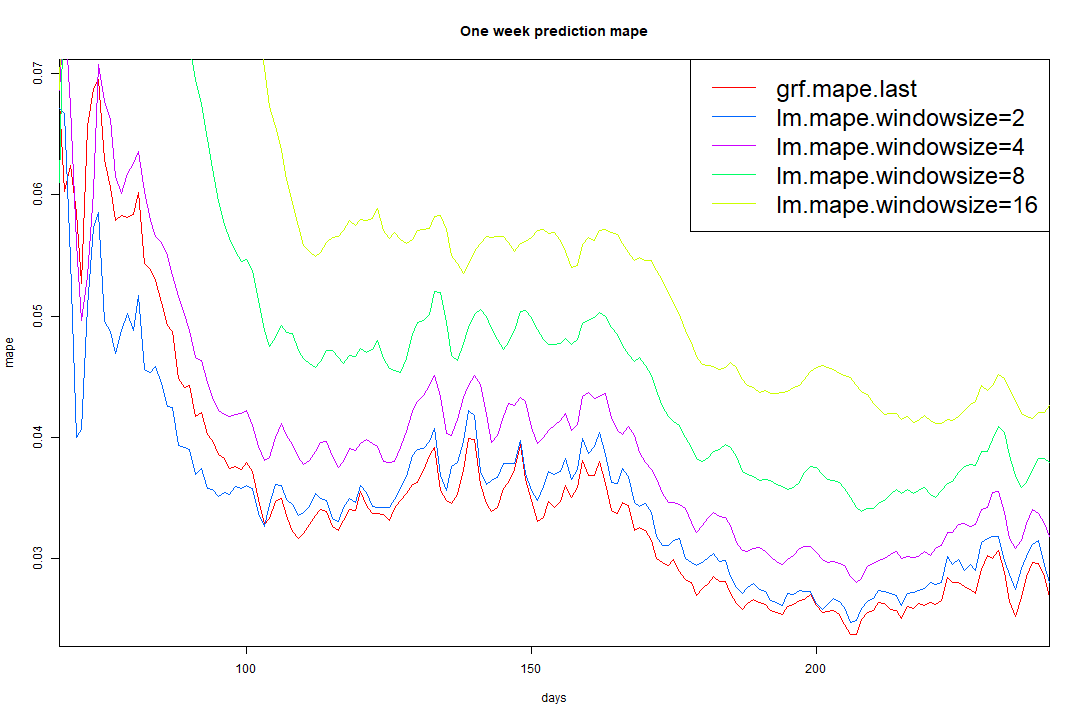}}
\end{figure}

\ \\

\newpage

\subsection{Median RMSEs and MAPEs}

\begin{table}[ht]
\floatconts
  {tab:3}%
  {\caption{Median RMSE and MAPE (3-day Moving Average)}}%
  {\begin{tabular}{lll}
  \toprule
  \bfseries Method & \bfseries RMSE & \bfseries MAPE\\
  \midrule
  OLS.wsize=16 & 0.3579 & 0.0584 \\
  OLS.wsize=8 & 0.3403  & 0.0532 \\
  OLS.wsize=4 &  0.3391 & 0.0510\\
  OLS.wsize=2 & 0.3447 & 0.0511\\
  \textbf{GRF} & \textbf{0.2913} & \textbf{0.0468}\\
  \bottomrule
  \end{tabular}}
\end{table}

\begin{table}[ht]
\floatconts
  {tab:4all}%
  {\caption{Median RMSE and MAPE (4-day Moving Average)}}%
  {\begin{tabular}{lll}
  \toprule
  \bfseries Method & \bfseries RMSE & \bfseries MAPE\\
  \midrule
  OLS.wsize=16 & 0.3574 & 0.0585 \\
  OLS.wsize=8 & 0.3374  & 0.0529 \\
  OLS.wsize=4 &  0.3120 & 0.0483\\
  OLS.wsize=2 & 0.3011 & 0.0455\\
  \textbf{GRF} & \textbf{0.2600} & \textbf{0.0423}\\
  \bottomrule
  \end{tabular}}
\end{table}

\begin{table}[ht]
\floatconts
  {tab:5}%
  {\caption{Median RMSE and MAPE (5-day Moving Average)}}%
  {\begin{tabular}{lll}
  \toprule
  \bfseries Method & \bfseries RMSE & \bfseries MAPE\\
  \midrule
  OLS.wsize=16 & 0.3499 & 0.0576 \\
  OLS.wsize=8 & 0.3256  & 0.0509 \\
  OLS.wsize=4 &  0.2867 & 0.0450\\
  OLS.wsize=2 & 0.2689 & 0.0410\\
  \textbf{GRF} & \textbf{0.2429} & \textbf{0.0395}\\
  \bottomrule
  \end{tabular}}
\end{table}

\ \\

\newpage

\begin{table}[ht]
\floatconts
  {tab:6}%
  {\caption{Median RMSE and MAPE (6-day Moving Average)}}%
  {\begin{tabular}{lll}
  \toprule
  \bfseries Method & \bfseries RMSE & \bfseries MAPE\\
  \midrule
  OLS.wsize=16 & 0.3426 & 0.0570 \\
  OLS.wsize=8 & 0.3120  & 0.0490 \\
  OLS.wsize=4 &  0.2672 & 0.0420\\
  OLS.wsize=2 & 0.2460 & 0.0378\\
  \textbf{GRF} & \textbf{0.2241} & \textbf{0.0361}\\
  \bottomrule
  \end{tabular}}
\end{table}

\begin{table}[ht]
\floatconts
  {tab:7}%
  {\caption{Median RMSE and MAPE (7-day Moving Average)}}%
  {\begin{tabular}{lll}
  \toprule
  \bfseries Method & \bfseries RMSE & \bfseries MAPE\\
  \midrule
  OLS.wsize=16 & 0.3342 & 0.0559 \\
  OLS.wsize=8 & 0.2956  & 0.0467 \\
  OLS.wsize=4 &  0.2480 & 0.0392\\
  OLS.wsize=2 & 0.2231 & 0.0346\\
  \textbf{GRF} & \textbf{0.2083} & \textbf{0.0336}\\
  \bottomrule
  \end{tabular}}
\end{table}

\ \\

\newpage

\subsection{RMSE Plots}

\begin{figure}[ht]
\floatconts
  {fig:RMSE3}
  {\caption{RMSE Plot (3-day Moving Average)}}
  {\includegraphics[width=\linewidth]{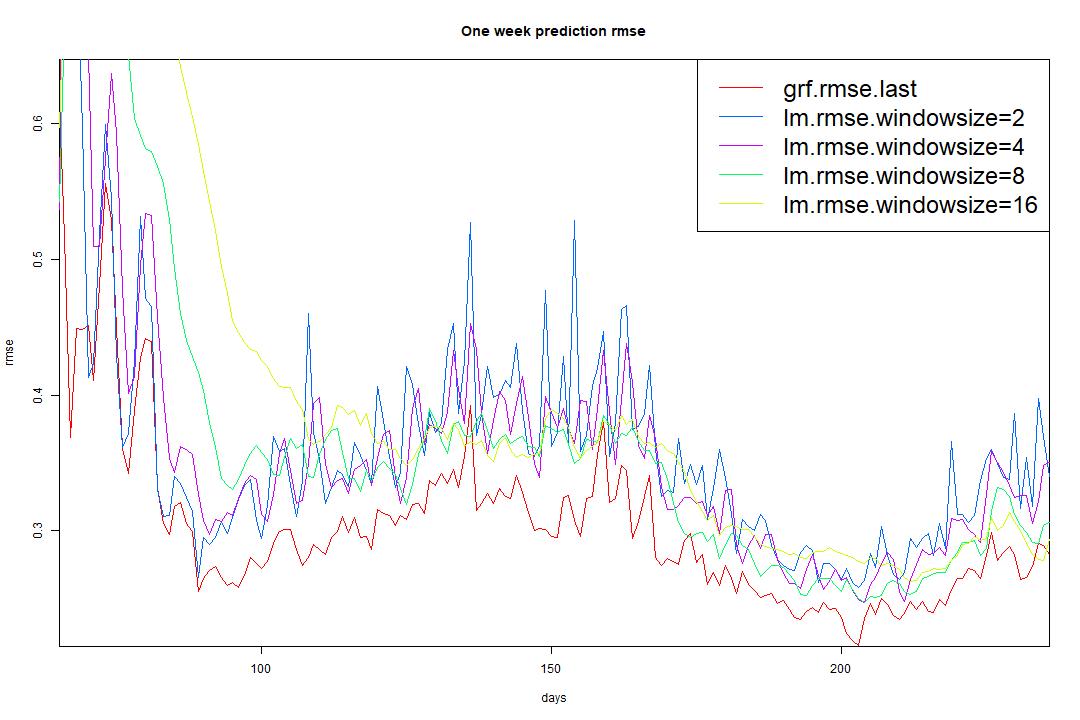}}
\end{figure}

\begin{figure}[t]
\floatconts
  {fig:RMSE4}
  {\caption{RMSE Plot (4-day Moving Average)}}
  {\includegraphics[width=\linewidth]{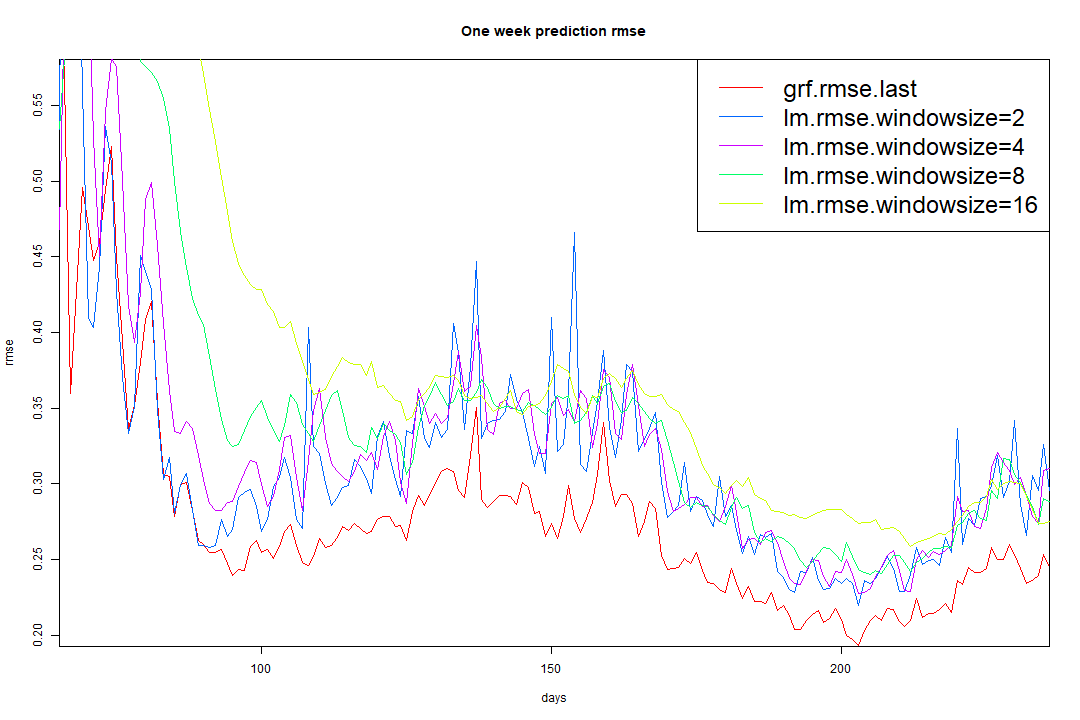}}
\end{figure}

\begin{figure}[t]
\floatconts
  {fig:RMSE5}
  {\caption{RMSE Plot (5-day Moving Average)}}
  {\includegraphics[width=\linewidth]{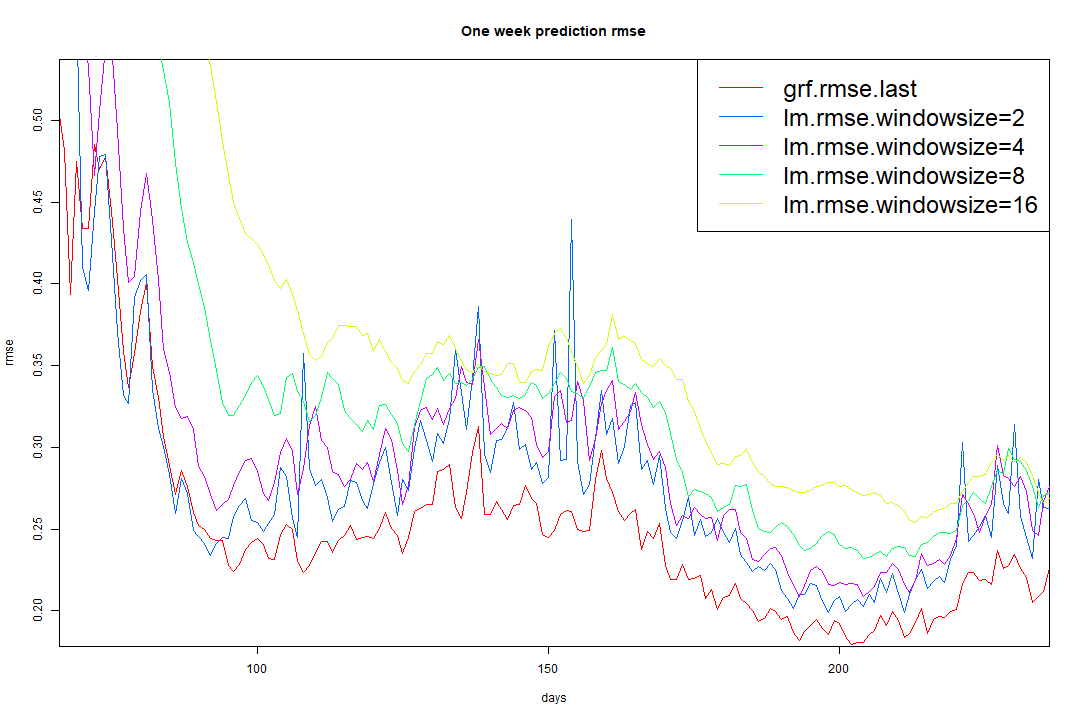}}
\end{figure}

\begin{figure}[ht]
\floatconts
  {fig:RMSE6}
  {\caption{RMSE Plot (6-day Moving Average)}}
  {\includegraphics[width=\linewidth]{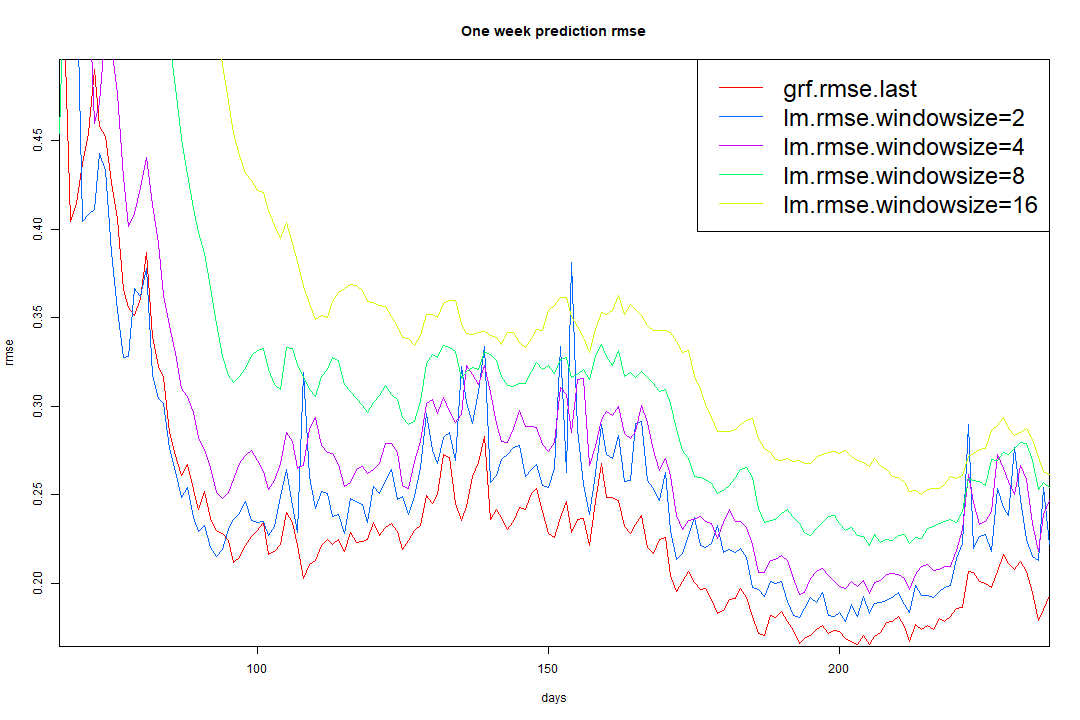}}
\end{figure}

\begin{figure}[ht]
\floatconts
  {fig:RMSE7}
  {\caption{RMSE Plot (7-day Moving Average)}}
  {\includegraphics[width=\linewidth]{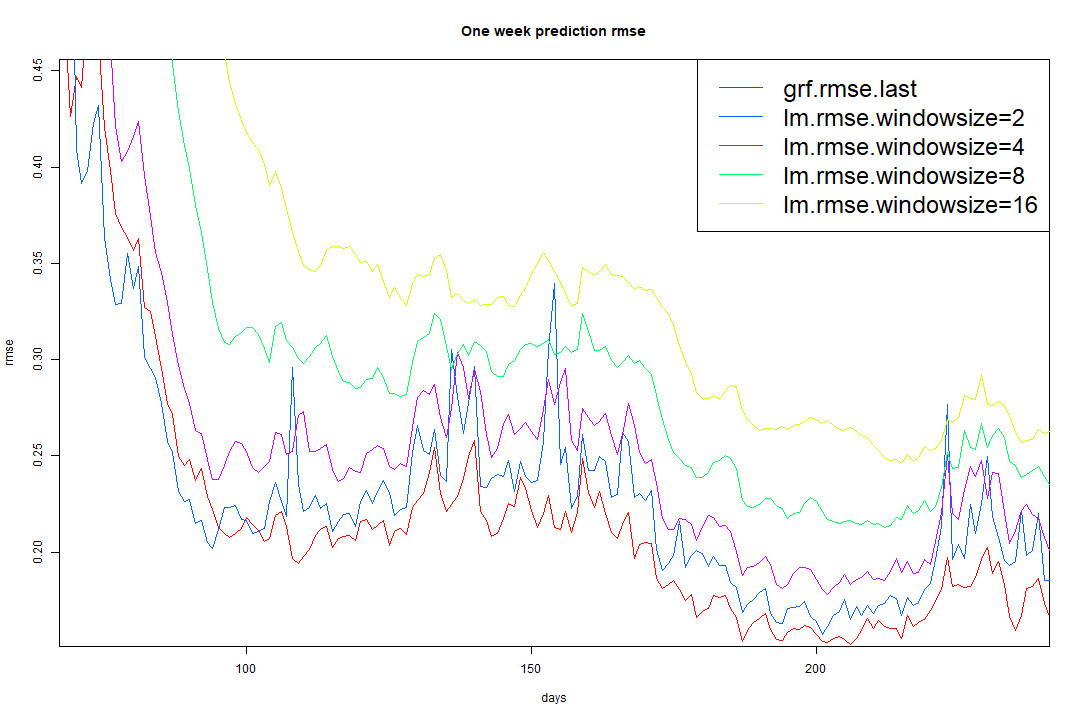}}
\end{figure}








\end{document}